\documentclass{article}

\usepackage{microtype}
\usepackage[T1]{fontenc}
\usepackage[utf8]{inputenc}
\usepackage{graphicx}
\usepackage{subcaption}
\usepackage{svg}
\usepackage{booktabs} 
\usepackage{multirow}
\usepackage{makecell}
\usepackage{wrapfig}
\usepackage{hyperref}

\usepackage[preprint]{icml2026}

\usepackage{amsmath}
\usepackage{amssymb}
\usepackage{mathtools}
\usepackage{amsthm}
\DeclareMathOperator*{\argtopk}{arg\,TopK}

\usepackage[most]{tcolorbox}
\usepackage{xcolor}
\usepackage{soul} 
\usepackage{siunitx}
\definecolor{softyellow}{RGB}{249, 236, 198}
\sethlcolor{softyellow}
\definecolor{lightyellow}{RGB}{255, 255, 204} 

\tcbset{
  myhighlight/.style={
    colback=lightyellow,      
    colframe=yellow!50!black, 
    boxrule=0.6pt,            
    arc=4pt,                  
    left=6pt, right=6pt, top=6pt, bottom=6pt, 
    enhanced,
  }
}
\newtcolorbox{highlightbox}{myhighlight}

\usepackage[capitalize,noabbrev]{cleveref}

\theoremstyle{plain}
\newtheorem{theorem}{Theorem}[section]

\theoremstyle{definition}
\newtheorem{definition}[theorem]{Definition}
\newtheorem{assumption}[theorem]{Assumption}
\theoremstyle{remark}

\usepackage[textsize=tiny]{todonotes}

\icmltitlerunning{Route Expert by Sequence, Not by Token}

\usepackage{amsmath,amsfonts,bm}

\def\Figref#1{Figure~\ref{#1}}

\def\Secref#1{Section~\ref{#1}}

\def\eqref#1{equation~\ref{#1}}

\def\1{\bm{1}}

\def\vh{{\bm{h}}}

\def\vs{{\bm{s}}}

\def\vx{{\bm{x}}}

\def\mS{{\bm{S}}}

\def\mW{{\bm{W}}}

\DeclareMathAlphabet{\mathsfit}{\encodingdefault}{\sfdefault}{m}{sl}
\SetMathAlphabet{\mathsfit}{bold}{\encodingdefault}{\sfdefault}{bx}{n}

\begin{document}

\twocolumn[
  \icmltitle{Route Expert by Sequence, Not by Token}


  \icmlsetsymbol{equal}{*}

  \begin{icmlauthorlist}
    \icmlauthor{Tiansheng Wen}{equal,sbu,xdu}
    \icmlauthor{Yifei Wang}{equal,mit}
    \icmlauthor{Aosong Feng}{yale}
    \icmlauthor{Long Ma}{xdu}
    \icmlauthor{Xinyang Liu}{ut}
    \icmlauthor{Yifan Wang}{sbu}
    \icmlauthor{Lixuan Guo}{xdu}
    \icmlauthor{Bo Chen}{xdu}
    \icmlauthor{Stefanie Jegelka}{tum,mit}
    \icmlauthor{Chenyu You}{sbu}
  \end{icmlauthorlist}
  \icmlaffiliation{sbu}{Stony Brook University}
    \icmlaffiliation{xdu}{Xidian University}
\icmlaffiliation{mit}{MIT}
\icmlaffiliation{yale}{Yale University}
\icmlaffiliation{ut}{UT Austin}  
\icmlaffiliation{tum}{TUM}  
\icmlcorrespondingauthor{Bo Chen}{ bchen@mail.xidian.edu.cn}
\icmlcorrespondingauthor{Chenyu You}{chenyu.you@stonybrook.edu}
  
  \icmlkeywords{Machine Learning, ICML}

{%
\renewcommand\twocolumn[1][]{#1}%
\begin{center}
    \centering
    \captionsetup{type=figure}
\includegraphics[width=1\textwidth]{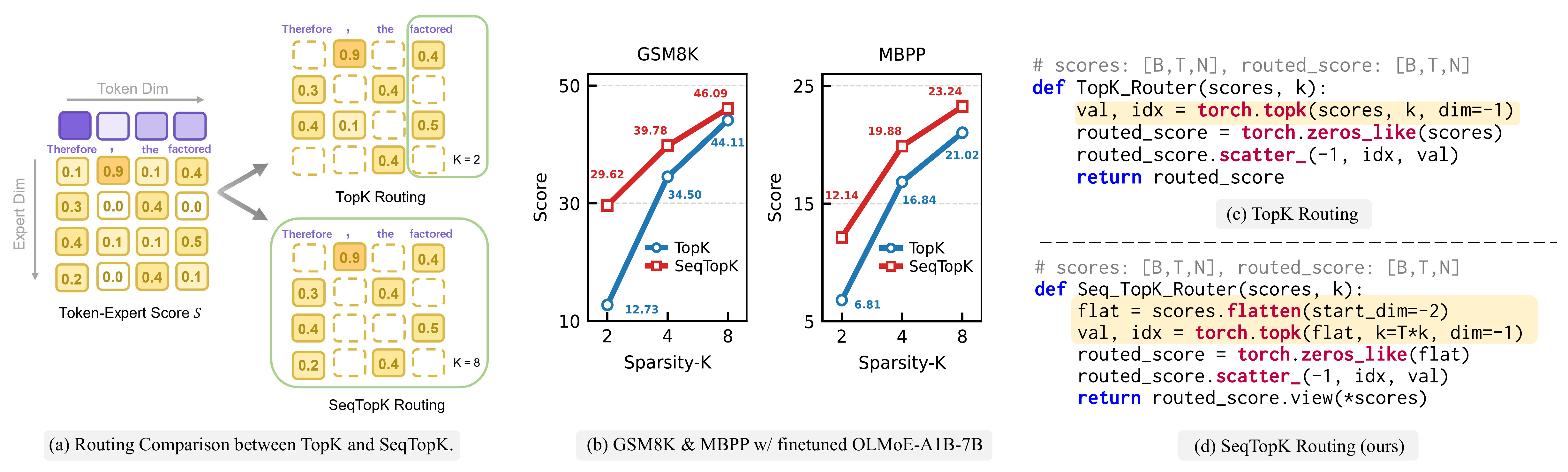}
    \captionof{figure}{
    \textbf{Overview of our proposed method.}
    \textbf{(a)} Illustrative comparison between standard TopK Routing and SeqTopK Routing in MoE models. 
    Under the same expert budget for a sequence ($T \cdot K = 3 \cdot 2 $ in this case), 
    SeqTopK routes experts via comparing expert scores across all tokens in a sequence, enabling dynamic and context-aware allocation of experts (e.g., more experts for hard tokens) via end-to-end training.
    \textbf{(b)} Performance of fintuned OLMoE-A1B-7B  on GSM8K and MBPP datasets. 
    SeqTopK consistently outperforms TopK under different expert budgets ($K = 2, 4, 8$), and the gain is much larger under sparser MoEs.
    \textbf{(c) \& (d)} Simple PyTorch implementations of TopK and SeqTopK routing. With a minimal modification of TopK routing (\hl{highlighted}), SeqTopK enables dynamic and context-aware expert allocation by comparing expert scores across all tokens in a sequence.
    }
    \label{fig:teaser}
    
\end{center}%
}

\vskip 0.3in
]



\printAffiliationsAndNotice{\icmlEqualContribution}

\begin{abstract}
Mixture-of-Experts (MoE) architectures scale large language models (LLMs) by activating only a subset of experts per token, but the standard \textit{TopK} routing assigns the same fixed number of experts to all tokens, ignoring their varying complexity. Prior adaptive routing methods introduce additional modules and hyperparameters, often requiring costly retraining from scratch. We propose \textbf{Sequence-level TopK (SeqTopK)}, a minimal modification that shifts the expert budget from the token level to the sequence level. By selecting the top $T \cdot K$ experts across all $T$ tokens, SeqTopK enables end-to-end learned dynamic allocation -- assigning more experts to difficult tokens and fewer to easy ones -- while preserving the same overall budget. SeqTopK requires only a few lines of code, adds less than 1\% overhead, and remains fully compatible with pretrained MoE models. Experiments across math, coding, law, and writing show consistent improvements over TopK and prior parameter-free adaptive methods, with gains that become substantially larger under higher sparsity (up to 16.9\%). These results highlight SeqTopK as a simple, efficient, and scalable routing strategy, particularly well-suited for the extreme sparsity regimes of next-generation LLMs.
Code is available at \href{https://github.com/Y-Research-SBU/SeqTopK}{here}.
\end{abstract}

\section{Introduction}
\begin{figure*}[t]
    \centering
\includegraphics[width=0.75\linewidth]{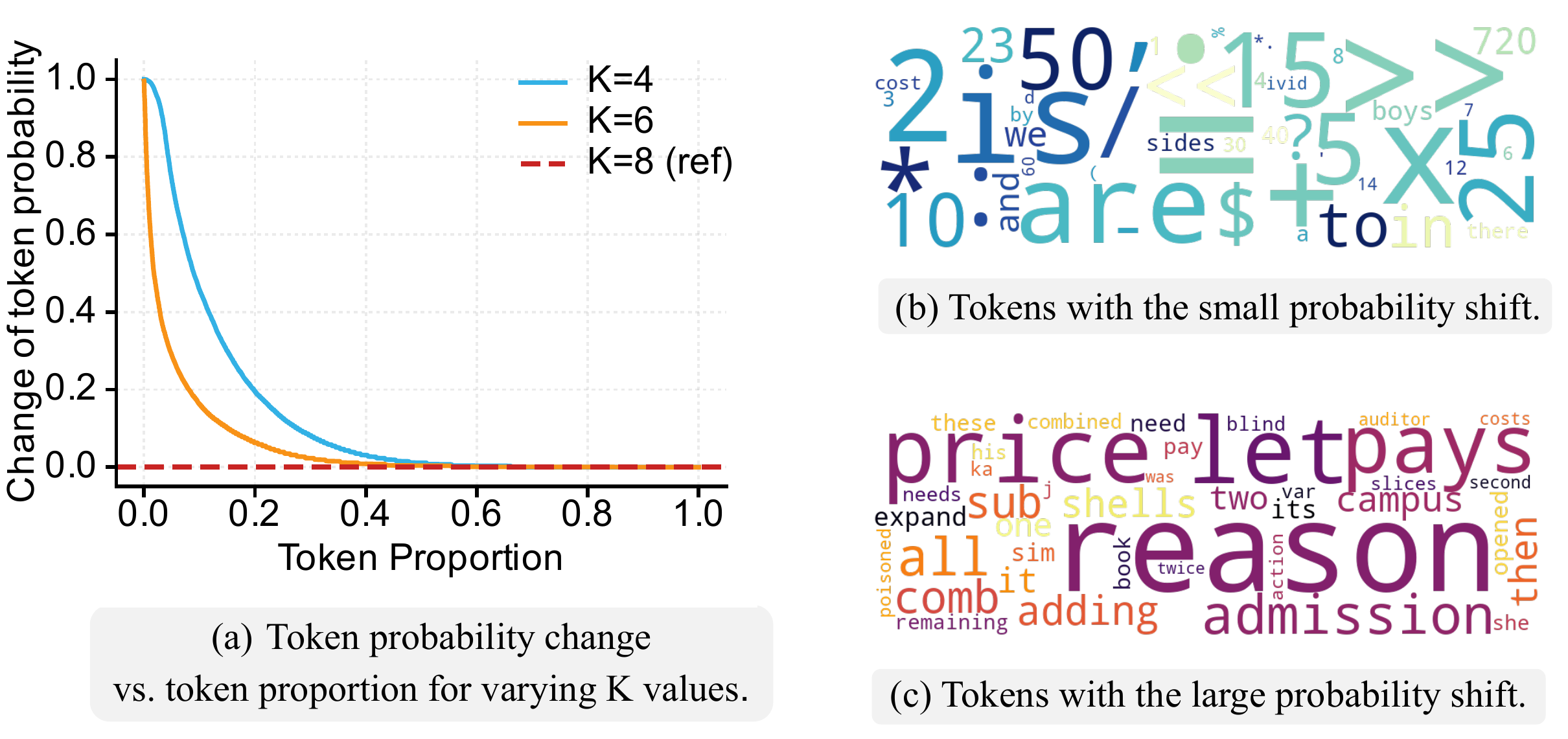}
    \caption{
    \textbf{Change of token probability $P(\vx_t|\vx_{<t})$ under varying active experts ($K$).}
    \textbf{(a) Token probability change vs. token proportion.} 
    Given the same prefix, we sample over 10k tokens across different K values, computing probability differences relative to $K=8$ (the original OLMoE setting). About 60\% of tokens show little probability change when $K$ is reduced from 8 to 4 or 6, while 10\% change dramatically, indicating that different tokens can require quite different numbers of activated experts to predict.
    \textbf{(b) \& (c) Word clouds of the top 50 tokens with small ($<0.01$) and large ($>0.5$)  token probability shifts.}
    Tokens with the larger probability differences are content words that influence semantic direction or topic shifts, whereas tokens with the smaller probability differences are numbers or function words that maintain structure. 
    }
    \label{fig:motivation}
    \vspace{-5pt}
\end{figure*}
Mixture-of-Experts (MoE) architectures have emerged as a central paradigm for scaling LLMs, offering massive capacity by activating only a subset of experts per token~\citep{shazeer2017outrageously,lepikhin2020gshard,fedus2022switch}.
However, the dominant \textit{TopK} routing assigns each token the same fixed number of experts ($K$), treating all tokens uniformly regardless of their predictive difficulty.

This uniformity overlooks the inherent heterogeneity of language~\citep{smith2013effect}. 
As shown in \Figref{fig:motivation}, 
we measure token probability change $|\Delta P(\vx_t|\vx_{<t})|$ when reducing $K$ from 8 (OLMoE's default) to 4 or 6.
Roughly 60\% of tokens are insensitive to expert budget changes, suggesting they can be predicted accurately with fewer experts, while around 10\% suffer substantial drops and clearly need more.
A uniform budget $K$ thus wastes computation on easy tokens while underserving hard ones.
Although prior adaptive routing methods~\citep{huang2024harder,lu2024not,wang2024remoe,guo2024dynamic,jin2024moe++} attempt to address this, they often require extra modules or pretraining from scratch, limiting compatibility with existing checkpoints.
We argue that these methods do not address the root cause: TopK's limited \emph{competition scope}. 
In fact, each token selects experts independently, with budgets decoupled across tokens -- there is no mechanism to \emph{reallocate} capacity from easy to hard tokens.

This raises a natural question: \emph{Can we enable such reallocation without introducing new modules? }

In this paper, we instead revisit standard TopK routing and propose a simple but effective relaxation: shifting the budget from the \emph{token level} to the \emph{context level}. 
Instead of enforcing $K$ experts per token, we allocate the top $T \cdot K$ experts across all $T$ tokens in a sequence. 
This \emph{Sequence-level TopK} (SeqTopK) routing enables the model, through end-to-end likelihood maximization, to assign more experts to challenging tokens and fewer to trivial ones, thereby achieving dynamic allocation without introducing new architectural components. 
Crucially, SeqTopK preserves the same overall budget as TopK, requires only a few lines of code to implement (Figure~\ref{fig:teaser}), and introduces no additional parameters or hyperparameters. It is fully compatible with existing MoE models and can be directly fine-tuned from pretrained TopK checkpoints. SeqTopK also integrates naturally with autoregressive decoding via an on-the-fly updated list of expert scores, which we term the \emph{Expert Cache} (analogous to the KV cache in attention). Overall, SeqTopK incurs less than 1\% overhead in computation and memory, while seamlessly extending the capabilities of existing MoE implementations.

Despite its simplicity, SeqTopK consistently outperforms standard TopK and prior adaptive methods. The gains are especially pronounced under higher sparsity (up to 16.89\%), highlighting its promise as state-of-the-art MoE models, such as GPT-OSS~\citep{agarwal2025gpt} and Qwen-Next~\citep{yang2025qwen3}, continue to scale with extreme sparsity. We validate SeqTopK across diverse domains, including math, coding, law, and writing, and demonstrate both its data-driven adaptive behavior that allocates more experts to harder tokens and more balanced expert utilization.

Our contributions are summarized as follows:
\begin{itemize}
    \item We propose \textbf{SeqTopK}, a routing mechanism that maintains the same compute budget as the TopK routing while enabling context-level dynamic expert allocation. 
    \item For efficient autoregressive inference, we introduce \textbf{online SeqTopK with Expert Cache}, which maintains an updated list of expert scores to support online routing that closely mirrors training-time SeqTopK.
    \item We provide extensive validation across multiple downstream tasks, showing that SeqTopK consistently outperforms the standard TopK and delivers larger gains as sparsity increases, along with qualitative evidence of its context-aware dynamic routing behavior and more balanced expert utilization.
\end{itemize}

\section{Background}
\textbf{Mixture of Experts (MoEs).} 
We first introduce a generic MoE architecture commonly used in Transformer-based language models (LMs) ~\citep{vaswani2017attention}.
A standard decoder-only transformer~\citep{radford2019language} comprises $L$ layers, where each block can be represented as follows:
\begin{align}
    \vx_t^l &= \text{Self-Attn}(\vh_{1:t}^{l-1}) + \vh_{1:t}^{l-1}, \\
    \vh_t^l &= \text{FFN}(\vx_t^l) + \vx_t^l, \quad t\in[T],
\end{align}
where $T$ denotes the length of the sequence $x_{1:T}$, $\text{Self-Attn}(\cdot)$ and $\text{FFN}(\cdot)$ are self-attention and the Feed-Forward Network.
Here, $\vx_t^l \in \mathbb{R}^{D}$ is the hidden state of the $t$-th token and $\vh_t^l \in \mathbb{R}^{D}$ the output of the $l$-th transformer block. We drop the layer index $l$ for better clarity.

A typical approach for constructing MoE language models involves substituting FFNs in transformers with MoE layers. 
These MoE layers partition the FFN into $N$ smaller FFNs, $E_1,\dots,E_N$, referred to as \emph{experts}, where $N$ denotes the total number of experts. Subsequently, each token $x_t$ is assigned to a few experts as determined by a router module $R(\cdot)$. The output is computed as: $\vh_t = \sum_{i=1}^N R_i(\vx_t) E_i(\vx_t) + \vx_t.$

\textbf{Standard TopK Routing in MoEs.}
TopK routing~\citep{shazeer2017outrageously,lepikhin2020gshard,fedus2022switch} is a widely used, de facto strategy for routing experts in modern LLMs.  
For each input token $x_t$, the router first computes the expert scores $\vs_t = \sigma(\mW\vx_t)$, where $\mW \in \mathbb{R}^{N \times D}$ and $\sigma(\cdot)$ denotes the \textit{softmax} function that normalizes the scores over $N$ experts. The router then identifies the $K$ most relevant experts and defines the routing gate $R_i(\vx_t)$ as:
\begin{flalign}
&\text{\footnotesize (TopK routing)} \nonumber \\
&R_i(\vx_t) =
\begin{cases} 
        1, & \text{if } i \in \argtopk_{i\in[N]}\ \vs_{t,i}\\
        0, & \text{otherwise} 
        \label{eq:topk}
\end{cases}
\end{flalign}
where 
$\argtopk(\cdot)$ returns the index set of the $k$ largest elements along the chosen dimension\footnote{Since the primary distinction between TopK and SeqTopK lies in expert allocation, we omit the final routing-weight selection step in Eq.~\ref{eq:topk} (\emph{i.e.}, selecting weights from $s_t$ based on the chosen indices) for simplicity.}.

Due to the limited space, we defer a detailed discussion on other related works to Appendix~\ref{sec:related}.

\section{Method}
\subsection{SeqTopK: From Token-level to Sequence-level Routing}
\label{sec:seqtopk}
\textbf{Token Heterogeneity in Standard TopK. }
While standard TopK routing adaptively selects experts for each token, it enforces a \emph{fixed budget of $K$ experts per token}, regardless of predictive difficulty. This assumption ignores the heterogeneity among tokens in a sequence. As shown in Figure~\ref{fig:motivation}, when reducing the routing budget from $K=8$ to $K=4$, over $60\%$ of tokens exhibit \emph{little or no degradation} in their predicted likelihood $P(\vx_t \mid \vx_{<t})$, whereas approximately $15\%$ of tokens suffer a \emph{drop of more than $0.5$}. 
{Theoretically, we prove in Section~\ref{sec:theoretical-proof} that such uniform budget allocation is inherently suboptimal.}
{Collectively, this evidence highlights substantial variation in token difficulty and the computational resources required.  
This suggests that standard TopK over-serves easy tokens and under-serves difficult ones, leading to inefficiency and degraded accuracy in sparse regimes. 
}

\textbf{Limitation of Token-level Routing.}
Several works attempt adaptive routing by replacing TopK with learnable thresholding functions (e.g., ReLU gating in ReMoE or identity experts~\citep{jin2024moe++,meituanlongcatteam2025longcatflashtechnicalreport}), thereby allowing different tokens to activate different numbers of experts. However, such methods show limited performance gains because they remain fundamentally at the local \emph{token level}: routing decisions depend only on each token’s own scores, without considering the surrounding context. In contrast, the power of self-attention in Transformers arises precisely from \emph{contextualization}: predictions are informed by comparing a token with its neighbors. MoE layers that treat tokens independently lack this context-aware adaptivity.

\textbf{Sequence-level Routing with SeqTopK. }
Motivated by these observations, we propose \textbf{SeqTopK}, a context-aware routing strategy that performs expert selection at the sequence level. Let
\begin{equation}
\begin{aligned}
    & \text{\footnotesize (SeqTopK scores)} \\
    & \mS = [\vs_1,\dots,\vs_T]^\top \in \mathbb{R}^{T\times N}, \vs_t = \sigma(\mW \vx_t) \in \mathbb{R}^N,
\end{aligned}
\end{equation}
where $\mS$ collects the expert scores for all tokens in a sequence of length $T$ with $N$ experts. SeqTopK selects top $K_{\text{seq}}$ entries across all $T\cdot N$ elements in the score matrix $\mS$:
\begin{flalign}
\label{eq:seqtopk}
&\text{\footnotesize (SeqTopK routing)} \nonumber \\
&R_i(\vx_t) =
    \begin{cases}
        1, & \text{if } (t,i) \in \argtopk_{t\in[T],\,i\in[N]} \mS_{t,i}, \\
        0, & \text{otherwise}. 
    \end{cases}
\end{flalign}

Here $K_{\text{seq}} = T \cdot K$, where $K$ is the original budget of TopK and $T$ is the number of tokens in a sequence, which ensures the total budget of SeqTopK exactly matches that of  TopK. Thus, with equal overall compute, tokens can now receive different numbers of experts. Empirically, SeqTopK allocates more experts to difficult tokens and fewer to trivial ones, improving performance without increasing cost. Importantly, the allocation depends not only on a token’s absolute score but also on its \emph{relative importance within the sequence}, thereby enabling context-aware routing in MoEs.

\textbf{Balanced Routing with Token-level Bounds.}
To prevent degenerate allocations (e.g., assigning no experts to some tokens), we enforce per-token bounds for the number of experts. Specifically, each token is guaranteed at least one expert, and capped at $K_{\text{tok}}+2$ experts, where $K_\text{tok}$ is the original token-level sparsity. 
This ensures all tokens contribute to MoE training while avoiding domination by a small subset of tokens. 
According to ablation studies in \Secref{sec:ablation-upper-bond}, we find that while SeqTopK's performance exhibits slight fluctuations depending on the specific upper bound chosen, the bounded variant consistently yields robust improvements across most of the evaluated settings.


\textbf{Compatibility with TopK MoE models.}
While many sophisticated designs for context-aware routing could be envisioned, we deliberately designed SeqTopK to remain closely aligned with standard TopK. At the implementation level, it essentially alters only the indexing dimensions of the TopK operator, \textit{requiring just a few lines of code to modify} (Figure~\ref{fig:teaser}(d)). This simplicity yields several practical advantages. First, no additional hyperparameters or auxiliary modules are introduced. Second, pretrained TopK MoE checkpoints can be readily adapted to SeqTopK with only a few hundred fine-tuning steps (Section~\ref{llm-ft}). 
Third, the computational and memory overhead introduced is negligible, typically less than $1\%$ in our profiling. Taken together, these properties make SeqTopK a practical, lightweight drop-in replacement for standard TopK routing, enabling MoEs to perform context-aware adaptive routing and deliver improved performance across diverse downstream tasks. 


\subsection{Online SeqTopK: Efficient Decoding with Cached Expert Scores}
\label{sec:online-sqtopk}

While SeqTopK seamlessly integrates into the training stage as a drop-in replacement for Top{K}, applying it directly at inference is problematic. Large language models generate tokens autoregressively from $t=1$ to $T$, yet vanilla SeqTopK requires access to the routing scores of all tokens $\vx_{1:T}$. In other words, vanilla SeqTopK is inherently \emph{non-causal}\footnote{Although non-causal, SeqTopK does not actually leak differentiable information about future tokens, since the $\arg\mathrm{TopK}$ operator in Eq.~\ref{eq:seqtopk} is non-differentiable.}, {and thus} makes it incompatible with autoregressive decoding where $\vx_{m+1:T}$ is {unavailable} at step $m$.
To address this, we propose Online SeqTopK supported by an efficient caching mechanism.

\textbf{Expert Cache.}
We first introduce an \textbf{Expert Cache} that stores all previously computed scores and is updated only with the scores at the latest step:
\begin{equation}
\begin{aligned}
    & \text{\footnotesize(Expert Cache)} \\
    & \mS_m =\begin{pmatrix}
    \mS_{m-1}\\
    \vs_m^\top
\end{pmatrix}, \quad 
\vs_m = \sigma(\mW \vx_m) \in \mathbb{R}^N.
\end{aligned}
\end{equation}
This design is analogous to the KV cache in self-attention: whereas the KV cache stores intermediate keys and values, the expert cache maintains routing scores for reuse. 
In practice, the expert cache can be co-located with the KV cache and managed with identical read/write mechanisms.
Its memory footprint is also modest: while the KV cache has size $B\times T \times H$ (with hidden dimension $H\approx 4096$), the expert cache is only $B\times T \times N$ (with $N\in [8,128]$ in typical MoE models). Consequently, Online SeqTopK introduces negligible memory and compute overhead.
\begin{table*}[t!]
\centering
\caption{
\textbf{Benchmark results of models fine-tuned from the OLMoE-A1B-7B model (64 routing experts) with different \textit{routing methods}.}
SeqTopK outperforms baseline methods across all settings, with gains becoming much larger (up to 7.55\%) as sparsity increases.  
}
\begin{tabular}{lcc|ccccc|c}
\toprule

\multicolumn{3}{l}{\textit{OLMoE-A1B-7B}}
& \textbf{GSM8k} & \textbf{MBPP} & \textbf{HumanEval} & \textbf{Summary} & \textbf{Law}  & \textbf{Avg} \\
{\textbf{Routing}}
& \multirow{2}{*}{\textbf{K}} & \textbf{Sparsity}&{0-shot} & {3-shot} & {0-shot} & {0-shot} & \small{0-shot} &-  \\

\textbf{Methods} & &\textbf{Ratio}& {EM} & {Pass@1} & {Pass@1} & {Score} & {Score} & Score   \\ \midrule
\addlinespace
Base
& {8}
& {1/8}
& 15.58 
& 19.80 
& 10.97 
& 7.49 
& 5.70 
& 11.91 \\
\midrule
TopK
& \multirow{4}{*}{8}
& \multirow{4}{*}{1/8}
& 44.11 
& 21.04 
& 13.41 
& 45.31 
& 24.89 
& 29.74  \\
MRL-TopK
&
&
& 43.93 
& 21.78 
& 12.21 
& 44.25 
& 21.08 
& 28.65  \\
BatchTopK
&
&
& 44.80 
& 22.63 
& 14.02 
& 42.89 
& 22.89 
& 29.45  \\
\textbf{SeqTopK}
&
&
& \colorbox{orange!10}{\textbf{46.09}}  
& \colorbox{orange!10}{\textbf{23.21}} 
& \colorbox{orange!10}{\textbf{15.24}} 
& \colorbox{orange!10}{\textbf{46.40}} 
& \colorbox{orange!10}{\textbf{26.52}}
& \colorbox{orange!10}{\textbf{31.49}} \\
\midrule
TopK
& \multirow{4}{*}{4}
& \multirow{4}{*}{1/16}
& 36.50 
& 16.84 
& 10.98 
& 41.42 
& 17.77 
& 24.70  \\
MRL-TopK
&
&
& 27.44 
& 17.82 
& 9.76 
& 39.89 
& 12.21 
& 21.42  \\
BatchTopK
&
&
& 35.93 
& 18.62 
& 12.82 
& 39.79 
& 17.82 
& 24.99  \\
\textbf{SeqTopK}
&
&
& \colorbox{orange!10}{\textbf{39.78}}      
& \colorbox{orange!10}{\textbf{19.88}}     
& \colorbox{orange!10}{\textbf{13.41}}      
& \colorbox{orange!10}{\textbf{43.00}}      
& \colorbox{orange!10}{\textbf{20.20}}    
& \colorbox{orange!10}{\textbf{27.25}}\\
\midrule
TopK                  
& \multirow{4}{*}{2}
& \multirow{4}{*}{1/32}
& 12.73      
& 6.81     
& 6.12          
& 37.89        
& 8.91    
& 14.49     \\
MRL-TopK              
& 
&
& 2.2      
& 4.2     
& 4.8          
& 19.6        
& 3.4    
& 6.84     \\
BatchTopK                       
& 
&     
& 18.49
& 6.60    
& 6.71                
& 37.18      
& 11.89   
& 16.17     \\
\textbf{SeqTopK}              
& 
&         
& \colorbox{orange!10}{\textbf{29.62}}      
& \colorbox{orange!10}{\textbf{12.14}}    
& \colorbox{orange!10}{\textbf{12.22}}      
& \colorbox{orange!10}{{\textbf{42.60}}  }    
& \colorbox{orange!10}{\textbf{14.10}}   
& \colorbox{orange!10}{\textbf{22.04}}      \\
\bottomrule
\end{tabular}
\label{tab:performance-olmoe}
\end{table*}
\begin{table*}[t]
\centering
\caption{
\textbf{Benchmark results of models fine-tuned from the Qwen1.5-MoE-A2.7B model with different \textit{routing methods}.}
This model by default, uses 4 shared experts and then selects 4 experts out of 60 routing experts. SeqTopK consistently outperforms the others.
}
\begin{tabular}{lcc|ccccc|c}
\toprule

\multicolumn{3}{l}{\textit{Qwen1.5-MoE-A2.7B}}
& \textbf{GSM8k} & \textbf{MBPP} & \textbf{HumanEval} & \textbf{Summary} & \textbf{Law}  & \textbf{Avg} \\
\multirow{2}{*}{\textbf{Methods}}
& \multirow{2}{*}{\textbf{K}} & \textbf{Sparsity-}&{0-shot} & {3-shot} & {0-shot} & {0-shot} & \small{0-shot} &-  \\
 & &\textbf{Ratio}& {EM} & {Pass@1} & {Pass@1} & {Score} & {Score} & Score   \\ \midrule
\addlinespace
Base 
& 4
& 1/8
& 38.69
&  {38.84}
& 32.31 
& 28.29
& 18.20 
& 31.27  \\
\midrule
TopK
& \multirow{4}{*}{4}
& \multirow{4}{*}{1/8}
& 55.16
& 35.02 
& 36.89 
& 39.21 
& 42.32 
& 41.72  \\
MRL-TopK
&
&
& 54.88
& 35.76
& 36.13
& 38.74
& 40.25
& 41.15 \\
BatchTopK
&
&
& 53.92
& 35.24
& 35.89 
& 37.10
& 42.39
& 40.91 \\
\textbf{SeqTopK}
&
&
&  \colorbox{orange!10}{\textbf{55.87}}  
& \colorbox{orange!10}{\textbf{36.41}}
&  \colorbox{orange!10}{\textbf{37.20}} 
&  \colorbox{orange!10}{\textbf{41.31}} 
&  \colorbox{orange!10}{\textbf{45.29}}
&  \colorbox{orange!10}{\textbf{43.19}} \\
\bottomrule
\end{tabular}
\label{tab:performance-qwen}
\vspace{-1mm}
\end{table*}
\begin{table*}[!]
\centering
\caption{\textbf{Zero-shot performance of 182M-sized MoE models pretrained with different \textit{routing methods}}, 
where ReMoE uses ReLU as an adaptive routing function.
SeqTopK shows best performance at each sparsity level; and comparing the two levels, SeqTopK excels at $K=4$ while others excel at $K=8$, indicating that SeqTopK benefits more from higher MoE sparsity. 
Best in  \colorbox{orange!10}{\textbf{bold}}.
}
\begin{tabular}{lcccccccc}
\toprule
Method   
& Experts & K &Sparsity-Ratio & LAMBDA & RACE & ARC-E & ARC-C & Avg. 
\\
\midrule
TopK   
&\multirow{3}{*}{128}            
        &\multirow{3}{*}{4}     
        &\multirow{3}{*}{1/32}
        &59.30        
        &27.33      
        &46.66       
        &20.73  
        &38.51     
        \\
ReMoE   
        &            
        &
        &
        &59.94        
        &28.01      
        &47.71       
        &21.07  
        &39.18     
        \\
\textbf{SeqTopK} 
         &            
         &   
         &
         &\colorbox{orange!10}{\textbf{60.82}}        
         &\colorbox{orange!10}{\textbf{28.70}}      
         &\colorbox{orange!10}{\textbf{48.07}}       
         &\colorbox{orange!10}{\textbf{23.41}}  
         &\colorbox{orange!10}{\textbf{40.25}}
         \\
\midrule
TopK   
        &\multirow{3}{*}{64}            
        &\multirow{3}{*}{8}  &\multirow{3}{*}{1/8}   
        &60.15        
        &27.61      
        &47.19       
        &20.06    
        &38.75 
        \\
ReMoE    &
         &            
         &         
         &\colorbox{orange!10}{\textbf{61.73}}       
         & 28.15     
         & \colorbox{orange!10}{\textbf{49.12}}      
         & 19.06 
         & 39.52 
         \\
\textbf{SeqTopK} &
         &            
         &          
         &60.67     
         &\colorbox{orange!10}{\textbf{28.55}}      
         &48.42     
         &\colorbox{orange!10}{\textbf{22.36}}
         &\colorbox{orange!10}{\textbf{40.00}} 
         \\

\bottomrule
\end{tabular}
\label{tab:pre-train-results}
\end{table*}

\textbf{Online SeqTopK.}
Based on the scores stored in the Expert Cache, we define the \textbf{Online SeqTopK}, which performs an expert competition against the historical scores in the cache at step $m$. 
The routing decision $R_i(x_m)$ is defined as:
\begin{equation}
\label{eq:online-seqtopk}
\begin{aligned}
    & \text{\footnotesize (online SeqTopK scores)} \\
    & \mS_m = [\mS_{m-1},\vs_m^\top] \in \mathbb{R}^{m\times N}, \quad \vs_m = \sigma(\mW \vx_m) \in \mathbb{R}^N \\[1ex]
    & \text{\footnotesize (online SeqTopK routing)} \\
    & R_i(\vx_m) =
    \begin{cases}
        1, & \text{if } (m,i) \in \argtopk_{t\in[m],\,i\in[N]} \mS_{t,i}, \\
        0, & \text{otherwise}.
    \end{cases}
\end{aligned}
\end{equation}
Here we set the budget to $K_{\text{seq}} = m \cdot K$, so that at most $m \cdot K$ experts are activated across the first $m$ tokens. 
When $m=T$, Online SeqTopK recovers the vanilla formulation in Eq.~\ref{eq:seqtopk}. 
While $x_m$ uses the cache to autonomously determine its own expert count, it cannot retrospectively alter the routing assignments of earlier tokens.
In other words, Eq.~\ref{eq:online-seqtopk} performs a virtual competition between the cumulative history and the current token, but the outcome only affects the routing allocations for the current step.
Nevertheless, Online SeqTopK guarantees that for any $1 \leq m \leq T$, the cumulative number of activated experts never exceeds $m \cdot K$, and is thus always bounded by the $T \cdot K$ experts of the vanilla formulation. 
This property establishes a natural upper bound on computation for efficient autoregressive decoding, while in expectation yielding the same total number of activated experts as the offline variant.

\subsection{Comparison with BatchTopK}
A closely related variant of SeqTopK is {BatchTopK}~\citep{bussmann2024batchtopk}, which selects experts across all tokens in a batch rather than within a single sequence. While effective in MoE-based diffusion image modeling, where batches are regular and homogeneous~\citep{yuan2025expert,shi2025diffmoe}, BatchTopK is ill-suited for language modeling. This is because text sequences in a batch vary in length and domain, making cross-sequence competition unfair and unstable. In deployment, batch sizes are flexible and unpredictable, creating mismatches between training and inference when routing depends on global batch composition. Moreover, selecting experts across independent user inputs risks information leakage, raising privacy and safety concerns in multi-tenant systems. 
In contrast, SeqTopK restricts competition to tokens within the same sequence, ensuring contextual consistency, robustness to batch size, and data privacy, making it better suited for adaptive routing in language modeling. 
We will show soon that it yields consistent gains over BatchTopK in practice with a further sensitivity study on batch size detailed in Appendix~\ref{sec:abl_bsz}.

\section{Benchmark Results and Analysis}
\label{sec:}
In this section, we evaluate our proposed SeqTopK routing strategy from both pre-training and fine-tuning perspectives.
\Secref{llm-ft} presents fine-tuning results for modern language MoE models under varying sparsity levels across tasks, including mathematical reasoning, code generation, summarization, and legal assistance.
\Secref{pre-train} reports GPT-2–level MoE pre-training results and zero-shot performance on benchmark datasets.
A detailed efficiency analysis is presented in \Secref{llm-cost}, examining the additional training and inference costs of SeqTopK.
\subsection{Scaling Model Sparsity at Fine-Tuning Stage}
\label{llm-ft}
\textbf{Experiment Setup.}
We select MRL-TopK~\citep{bussmann2025learning}, BatchTopK~\citep{bussmann2024batchtopk}, and standard TopK as baselines because they \textbf{enable MoE model fine-tuning} while \textbf{only varying in their routing mechanism} and requiring \textbf{no additional parameters.}
We fine-tuned two models~\citep{muennighoff2024olmoe,qwen_moe} and evaluated them on five downstream datasets using the framework of~\citet{feng2023ernie}. 
We then scaled the models to extreme sparsity (\emph{e.g.}, reducing the number of activated experts per token from 8 to 2) and reported performance across different sparsity levels.  
Zero-shot evaluations were conducted on GSM8K~\citep{cobbe2021training}, HumanEval~\citep{chen2021evaluating}, Summary~\citep{wang2024letexpertsticklast}, and Law~\citep{wang2024letexpertsticklast}, while 3-shot evaluations were performed on MBPP~\citep{austin2021program}. 
Note that BatchTopK is sensitive to the evaluation batch size (see Appendix~\ref{sec:abl_bsz}). 
Therefore, we report the best performance in Table~\ref{tab:performance-olmoe} and Table~\ref{tab:performance-qwen}.
Detailed settings are provided in \Secref{ft-setting}.  

\textbf{Analysis.} Tables~\ref{tab:performance-olmoe} and \ref{tab:performance-qwen} present a comparative analysis of OLMoE-A1B-7B and Qwen1.5-MoE-A2.7B across five downstream tasks under varying routing regimes. Evaluation encompasses the original ``Base" models alongside those fine-tuned using standard \textit{TopK} routing and the proposed \textit{SeqTopK} strategy.
Under default sparsity configurations (activating 8/64 experts for OLMoE and 4/60 for Qwen1.5), SeqTopK yields average performance improvements of \textbf{5.9\%} and \textbf{3.6\%}, respectively. 
Unlike existing dynamic routing methods that necessitate training from scratch \citep{jin2024moe++, guo2024dynamic}, SeqTopK is seamlessly compatible with fine-tuning. While prior approaches often underperform during fine-tuning due to architectural discrepancies, SeqTopK provides a low-overhead solution that readily adapts to pre-trained checkpoints. As detailed in \Secref{sec:comp_moe++}, SeqTopK yields substantial gains over MoE++, demonstrating that extending the routing scope from token-level to sentence-level effectively enables context-aware expert allocation without the need for intensive re-training.

\textbf{Analysis of scaling sparsity results.}
Table~\ref{tab:performance-olmoe} compares routing strategies across varying sparsity levels. MRL-TopK exhibits marked degradation as sparsity increases, while BatchTopK—though competitive—remains sensitive to evaluation batch size, limiting its deployment. 
In contrast, SeqTopK consistently outperforms all baselines, with the performance gap widening as sparsity increases (i.e., $K$ decreases from 8 to 2). 
In the extreme $K=2$ regime, SeqTopK achieves nearly twice the performance of standard TopK. 
This suggests that SeqTopK’s context-aware allocation is particularly advantageous for ultra-sparse models, where maximizing limited expert capacity is critical, making it highly effective for both pre-training and fine-tuning.

\subsection{GPT2-level Pre-training evaluation}
\label{pre-train}
\textbf{Experiment Setup.}
We select ReMoE~\citep{wang2024remoe} (ReLU-routed) and standard TopK serve as our baselines, distinguished \textbf{only by their routing mechanism}.
Other related work, such as MoE++~\citep{jin2024moe++} and DynMoE~\citep{guo2024dynamic}, that incorporate additional zero or gate experts are excluded. 
We train the models on The Pile~\citep{gao2020pile} for 60k steps ($\sim$30B tokens) and
evaluate their zero-shot performance across three downstream tasks: LAMBADA~\citep{paperno2016lambada}; RACE~\citep{lai2017race};
ARC~\citep{clark2018think}.
A more detailed experiment
setup is provided in \Secref{pre-train-setting}.
\begin{figure*}
    \centering
    \includegraphics[width=1\linewidth]{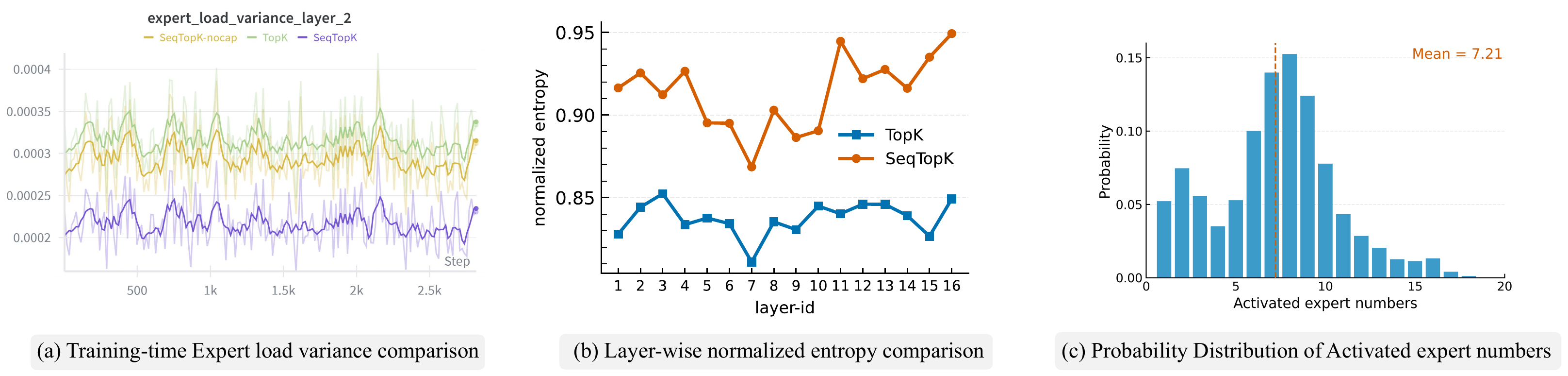}
    \caption{\textbf{Routing Dynamics of SeqTopK.} 
\textbf{(a) Expert load variance during training.}
Expert load variance was investigated at layer 2 during OLMoE fine-tuning on GSM8K. 
SeqTopK with an upper bound consistently delivered the lowest load variance. 
In comparison, SeqTopK without a cap showed a slightly lower variance than TopK, and crucially, it remains robust without leading to training instability or severe load imbalance.
    \textbf{(b) Layer-wise normalized entropy comparison. }
\textit{Higher} entropy means more \textit{balanced} expert utilization.
SeqTopK consistently exhibits higher entropy than TopK, suggesting that its sequence-level routing encourages more uniformed (\emph{i.e.}, balanced) expert utilization.
\textbf{(c) Expert Activation Pattern of online SeqTopK (w/o token-level upper bound).}
    We present the distribution of OLMoE's expert activation patterns on the GSM8K dataset with the SeqTopK routing strategy.
    SeqTopK learns to adaptively activate each token with a varying number of tokens from $1$ (lower bound) to $18$, following a normal-like distribution. 
}
    \label{fig:router_dynamics}
\end{figure*}

\textbf{Analysis.} 
Table~\ref{tab:pre-train-results} reports the zero-shot accuracy of different routing methods on downstream tasks across sparsity levels.
SeqTopK outperforms the baselines, achieving a notable gain of \textbf{3.2\%} and \textbf{4.5\%} over TopK across sparsity regimes with only 1\% additional overhead.
As sparsity increases, the performance margin widens, indicating that SeqTopK better handles high-sparsity settings through its sequence-level routing.
Compared with ReMoE~\citep{wang2024remoe}, which employs a three-stage training pipeline and is sensitive to sparsity-controlling hyperparameters, SeqTopK integrates seamlessly with existing MoE models, enabling dynamic routing with only a few lines of code.
\begin{table}[!]
    \centering
    \caption{\textbf{Efficiency Analysis for SeqTopK.} (a) GPU hours for pre-training a 182M-size model over 30B tokens using various routing strategies.
    (b) Decoding efficiency for online SeqTopK and the overhead introduced by Expert-Cache on   OLMoE, using \textit{huggingface.gen()} for fair comparison. }
    \label{tab:efficiency-all}
    \begin{subtable}{\columnwidth}
        \centering
        \caption{Pre-training Time}
        \label{tab:pretrain-throughput}
        \small 
        \begin{tabular}{l c}
            \toprule
            Method & \makecell{Wall-clock Time (GPU hours)} \\
            \midrule
            TopK             & 244.2 \\
            ReMoE            & 245.0 \\
            \textbf{SeqTopK} & 246.4 \textbf{(+1\%)} \\
            \bottomrule
        \end{tabular}
    \end{subtable}

    \vspace{4mm} 

    \begin{subtable}{\columnwidth}
        \centering
        \caption{Inference throughput and peak memory}
        \label{tab:inference-efficiency}
        \small
        \begin{tabular}{l c c}
            \toprule
            Method & Tokens/s & \makecell{Peak Memory (GB)} \\
            \midrule
            TopK             & 141.23 & 18.21 \\
            \textbf{SeqTopK} & 139.41 \textbf{(-1\%)} & 18.35 \textbf{(+1\%)} \\
            \bottomrule
        \end{tabular}
    \end{subtable}
    \vspace{-5pt}
\end{table}
\subsection{Efficiency analysis}
\label{llm-cost}
Table~\ref{tab:pretrain-throughput} details the GPU hours required for pre-training a 182M-parameter model on 30B tokens across various routing strategies. SeqTopK introduces a marginal pre-training overhead of only 0.9\% relative to vanilla TopK, significantly improving the efficiency–performance trade-off. We also benchmark decoding efficiency for online SeqTopK and the overhead of Expert-Cache using the \textit{huggingface.generate()} framework with KV-Cache enabled. As shown in Table~\ref{tab:inference-efficiency}, online SeqTopK maintains throughput nearly identical to standard TopK. Furthermore, Expert-Cache incurs a negligible 0.8\% increase in peak GPU memory usage; as analyzed in \Secref{sec:online-sqtopk}, its theoretical cost remains substantially lower than that of KV-Cache and requires only minor architectural modifications for integration.
\subsection{Empirical Understandings}
\label{sec:diff-topk}
In this section, we further provide an in-depth understanding of SeqTopK regarding its adaptive behaviors, the distribution of activated experts, and sensitivity w.r.t.~batch size and sequence length.

\begin{figure*}[t]
    \centering
\includegraphics[width=0.85\linewidth]{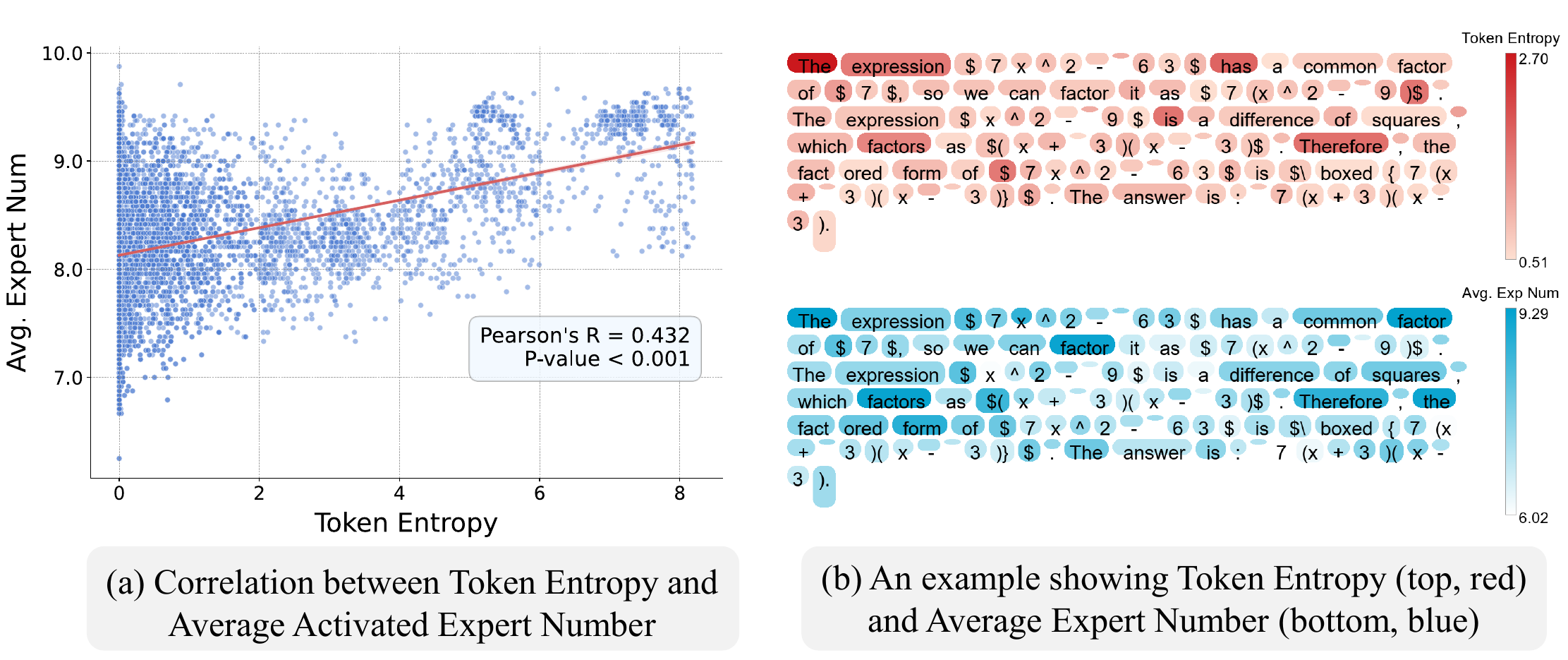}
\caption{\textbf{Correlation between token entropy and expert activation.} 
(a) Analysis of 10K tokens generated by fine-tuned Qwen-1.5-A2.7B, showing that higher token entropy—defined as the entropy of the output distribution at each token, capturing prediction uncertainty and token hardness—correlates well with a larger number of average activated experts. 
(b) Illustration from a specific generated sequence, where SeqTopK often activates more experts on high-entropy tokens, such as the ``\$'' symbol marking the start of a math expression.} 
\vspace{-5pt}
    \label{fig:entropy}
\end{figure*}

{\textbf{SeqTopK Enables More Balanced Expert Utilization.}
To assess the impact of SeqTopK on training stability, we evaluate expert load variance and routing entropy during OLMoE fine-tuning on GSM8K. 
As illustrated in \Figref{fig:router_dynamics}(a), SeqTopK with an upper bound consistently maintains lower load variance across all layers; notably, even its unconstrained variant outperforms the standard TopK baseline. This empirical evidence suggests that SeqTopK is intrinsically robust, ensuring stability without relying on rigid per-token caps.
We further analyze routing behavior by computing the normalized routing entropy, defined as $H = - \sum_{e=1}^E p_e\log{p_e} / \log{E}$, where $p_e$ is the mean assignment probability for expert $e$~\citep{lepikhin2020gshard,wu2024gw}.
A \textit{higher} entropy means a more \textit{balanced} expert utilization.
As shown in \Figref{fig:router_dynamics}(b), SeqTopK achieves higher entropy by generating softer, more distributed assignment probabilities. By adaptively allocating experts based on token importance, SeqTopK balances utilization across sequences, resulting in a more refined and robust capacity distribution than traditional TopK routing.
More visualizations can be found in Section \ref{sec:vis}.

\textbf{Normal-like Distribution of Activated Experts.}
To further elucidate the operational differences between SeqTopK and TopK, we visualize the expert activation patterns in \Figref{fig:router_dynamics}(a). We analyze the distribution of activated experts per token across the $16$ layers of an OLMoE model using 10K tokens from the GSM8K dataset. Unlike the fixed TopK baseline ($K=8$), SeqTopK (without token-level constraints) dynamically allocates between $1$ and $18$ experts per token, following a near-normal distribution. This behavior demonstrates an emergent adaptiveness: by autonomously determining the optimal expert budget for each token at every layer, SeqTopK achieves superior resource allocation and measurable performance gains.

\textbf{SeqTopK Activates More Experts on Hard Tokens.}
Inspired by~\citet{wang2025beyond}, we study how token entropy correlates with the average number of activated experts.
The token entropy for token $t$ is calculated by:
$H_t=-\sum_{j=1}^V p_{t,j}log(p_{t,j})$, where higher values signify greater predictive uncertainty and a need for increased computational resources.
\Figref{fig:entropy} confirms that the SeqTopK's context-aware design adaptively distributes the expert budget based on token difficulty.
High-entropy tokens that steer reasoning, such as ``therefore," receive a higher expert count, whereas low-entropy tokens representing routine steps or symbols, such as ``+", are assigned fewer. Beyond entropy-driven allocation, SeqTopK prioritizes key content words (\emph{e.g.}, ``factor") and structural markers like the ``\$" symbol, which denotes the initiation of mathematical expressions.
\begin{figure}[t]
    \centering
    \includegraphics[width=0.7\linewidth]{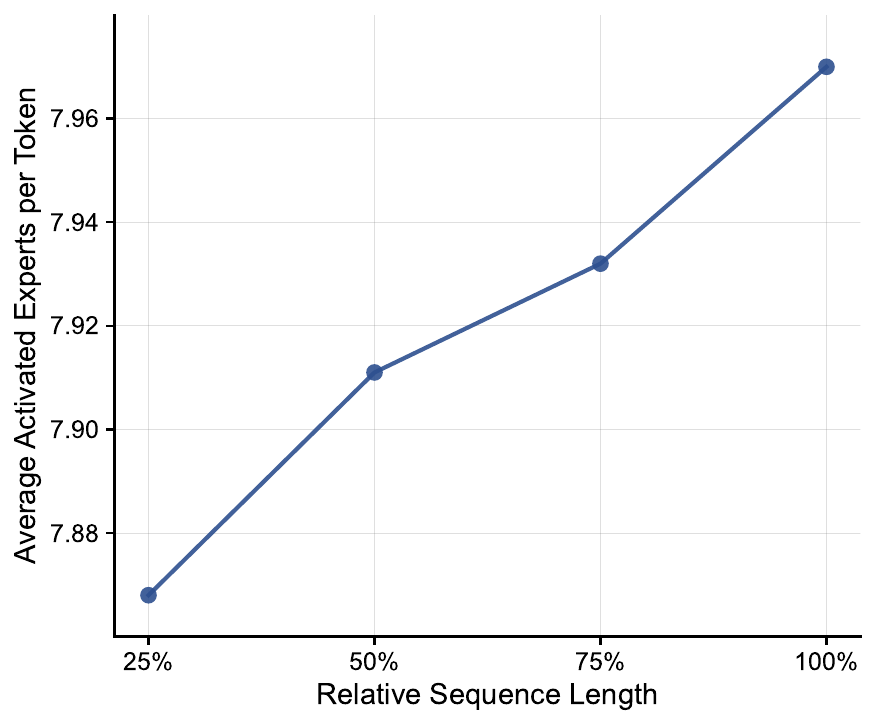}
    \caption{
    \textbf{Online SeqTopK activation patterns on GSM8K.}
    Average activated experts \textit{increase} with sequence length, demonstrating adaptive resource reallocation toward context-rich tokens.}
    \label{fig:online-pattern}
\end{figure}

\textbf{Online SeqTopK Learns Adaptive Banking Mechanism.}
To address whether the training-inference mismatch leads early tokens to over-consume experts due to lack of future competition, we analyze how expert allocation evolves during autoregressive generation on GSM8K sequences.
As shown in Figure~\ref{fig:online-pattern}, activated experts increase monotonically with sequence position, revealing a learned \textit{banking mechanism} where the model conserves budget on early tokens and reallocates surplus to later reasoning steps.
This adaptive pattern arises through end-to-end training, indicating the model learns to account for resource scarcity.
\begin{table}[t]
    \centering
    \caption{
    \textbf{Performance vs. Inference Answer Length of fine-tined OLMoE on GSM8K.}
    }
    \resizebox{\linewidth}{!}{
    \begin{tabular}{c|ccccc}
        \toprule
         Answer length&[0,50) &[50,100) & [100,150)& [150,250) &[250,)   \\
         \midrule
         TopK&60.48 &51.78 & 33.71& \textbf{25.08} &0.0 \\
         SeqTopK&\textbf{63.72} &\textbf{52.99} &\textbf{35.41}& 25.04 &0.0 \\
         \bottomrule
    \end{tabular}
}
    \label{tab:perf-answer-len}
\end{table}

\textbf{Sensitivity w.r.t. Sequence Length.}
To ensure dynamic routing does not introduce instability, we evaluate robustness to varying sequence lengths during inference. 
Results in Table~\ref{tab:perf-answer-len} show that SeqTopK introduces no additional fragility, consistently outperforming TopK across all intervals -- including a $+2.8\%$ gain in the $[100, 250)$ token range.
While accuracy for both methods declines as reasoning complexity increases with length, SeqTopK’s degradation closely tracks the baseline. 
This suggests that performance sensitivity stems from the pre-trained model's intrinsic limits rather than the routing strategy, confirming SeqTopK's robustness across diverse contexts.

\section{Conclusion \& Discussion}
We introduce Sequence-level TopK, a parameter-free replacement for TopK routing that adaptively redistributes expert capacity within a fixed sequence-level budget. 
By prioritizing informative tokens over trivial ones, SeqTopK optimizes model width via token-level specialization. This approach is orthogonal to pruning (length), early exiting (depth), and speculative decoding (generation), making it seamlessly compatible with existing efficiency frameworks for adaptive resource allocation.
\section*{Impact Statement}
This paper presents work whose goal is to advance the field of Machine
Learning. There are many potential societal consequences of our work, none
which we feel must be specifically highlighted here.
\nocite{langley00}

\bibliography{example_paper}
\bibliographystyle{icml2026}

\newpage
\appendix
\onecolumn
\section{Related Work}
\label{sec:related}
\textbf{Mixture-of-Experts.}
{The concept of Mixture-of-Experts (MoE) was first introduced by~\citet{jacobs1991adaptive,jordan1994hierarchical} as a way to model heterogeneous data with specialized modules.  
In this framework, only a subset of parameters, called \emph{experts}, is activated for each input, allowing the model to scale capacity without proportionally increasing computation.  
\citet{shazeer2017outrageously} extended this idea to large-scale language modeling, training LSTM-based MoE models that achieved performance competitive with dense counterparts.}
Early MoE works~\citep{roller2021hash,dai2022stablemoe} adopt fixed routing strategies, in which each expert is typically assigned a specific role to ensure stable routing and training.  
Subsequently, GShard~\citep{lepikhin2020gshard} and Switch Transformer~\citep{fedus2022switch} introduced learnable Top-K routing strategies, enabling MoE language models to scale to unprecedented sizes. However, these methods often suffer from load imbalance across experts, requiring auxiliary balancing losses~\citep{wang2024auxiliary}.  
{Building on this line,}
\citet{dai2024deepseekmoe} {proposed} fine-grained experts that increase the total number of experts by decomposing large experts into smaller ones ({e.g.}, splitting $8$ experts into $64$ experts with $0.25\times$ size each), thereby substantially improving combinatorial flexibility and enabling more precise specialization.
%
\citet{liu2024deepseekv2,liu2024deepseek} further introduce shared experts to capture and consolidate common knowledge across varying contexts and incorporate an expert-grouping strategy to enhance expert specialization. 
Beyond advances in model architectures and training strategies, 
{MoE concepts have been broadly adopted in diverse domains, including large language models~\citep{jiang2024mixtral,muennighoff2024olmoe,guo2025deepseek,li2025minimax,yang2025qwen3,team2025kimi}, multimodal LLMs~\citep{aria,lin2024moe,li2025uni,wu2024deepseekvl2mixtureofexpertsvisionlanguagemodels}, diffusion models~\citep{balaji2022ediff,feng2023ernie,sun2024ec}, and attention mechanisms~\citep{shen2024jetmoe,jin2024moh,fu2024moa,pikekos2025mixture}.}

Despite these advances, recent studies suggest that \textit{ultra-sparse} MoE architectures can offer superior efficiency with competitive performance.
For example, Qwen3-Next~\citep{yang2025qwen3} introduces an ultra-sparse MoE with only 10 activated experts per token out of 512 total experts, achieving competitive performance with dense models while incurring less than one-tenth of the training cost of dense models.
However, our experiments, as shown in \Figref{fig:teaser} (c), show that fixed Top-K routing struggles in such regimes, as the constant number of activated experts per token limits performance under extreme sparsity regime.  

To address this limitation, we propose SeqTopK, which allocates each token’s expert budget via sequence-level comparison, assigning more experts to high-entropy tokens and fewer to low-entropy ones for adaptive routing in ultra-sparse regimes.

\textbf{Adaptive Token Computation.}
{A central idea in efficient model design is to allocate computation adaptively based on token importance. This principle has been applied from compact representations~\citep{kusupati2022matryoshka,wenbeyond} to large language models~\citep{hou2020dynabert,schuster2022confident,dehghani2018universal}, consistently showing that selective computation can reduce cost while preserving, or even improving, generalization.    
Within the MoE framework, several works explore token-level dynamic routing.   \citet{huang2024harder,lu2024not} propose Top-P selection, where a token activates experts until their cumulative probability exceeds a threshold $p$.   ReMoE~\citep{wang2024remoe} replaces TopK and softmax with ReLU gating, coupled with $L_1$ regularization to encourage sparsity and balance.
Other methods take a global perspective: \citet{guo2024dynamic,wu2025grove} add a learnable threshold or a judging expert to enforce cross-token adaptivity.
\citet{jin2024moe++,meituanlongcatteam2025longcatflashtechnicalreport} add zero-compute experts, allowing tokens routed to them to bypass to the next layer, enabling a dynamic per-token budget.
Beyond horizontal expert routing, MoD ~\citep{raposo2024mixture} and MoR~\citep{bae2025mixture} have treated adaptive token computation as a depth allocation problem and route tokens vertically to dynamically control computation depth.
}

Despite these advances, existing approaches face several critical limitations that hinder their widespread adoption. 
They typically rely on empirical thresholds that generalize poorly across diverse scenarios~\citep{huang2024harder,lu2024not}, introduce additional gating parameters that complicate integration with established frameworks~\citep{guo2024dynamic,wu2025grove}, and employ multi-stage training procedures that substantially increase computational complexity while undermining training stability~\citep{wang2024remoe}.
Overall, these methods realize dynamic budgeting via \textit{expert reduction}; in contrast, SeqTopK offers a \textit{bidirectional} choice for each token, increasing or decreasing token budgets as needed.
SeqTopK neither requires auxiliary parameters nor imposes additional training overhead, offering a streamlined solution that integrates seamlessly into existing architectures for efficient fine-tuning and continued pre-training.

\section{Theoretical Insight}
\label{sec:theoretical-proof}

In this section, we provide theoretical insights into the superiority of {SeqTopK} over {TopK} by framing the routing mechanism as a \textit{constrained resource allocation problem}. 
We posit that the core advantage of SeqTopK lies in its ability to exploit the inherent \textbf{heterogeneity} of input instances. By modeling the number of activated features (experts) as a limited computational budget, we demonstrate that an adaptive strategy (corresponding to SeqTopK) achieves optimality by adhering to the \textit{equimarginal principle~\citep{marshall2013principles}}, dynamically shifting resources from ``easy" samples to ``hard" samples.

\subsection{Preliminaries}

Consider a dataset $\mathcal{D} = \{(x_i, y_i)\}_{i=1}^N$ with $N$ data points. 
To gain theoretical insight without loss of generality, we analyze a simplified setting where the \textit{resource is modeled as the continuous number of features} utilized for each instance.
Let $m_i \in \mathbb{R}_{\ge 0}$ denote the number of features allocated to the $i$-th data point and $\ell_i(m_i)$ represent the loss function for the $i$-th instance given $m_i$ features (e.g., the negative log-likelihood in logistic regression).
The total feature budget is fixed at $B$.
We assume a continuous relaxation of $m_i$ for gradient-based analysis.

Following~\citet{marshall2013principles}, we establish the standard assumption of the principle of diminishing marginal returns in Assumption~\ref{ass:convexity}.
Based on this, we define two resource allocation strategies that satisfy the budget constraint $\sum_{i=1}^N m_i = B$: \textbf{uniform} (i.e., TopK routing) and \textbf{adaptive} (i.e., SeqTopK routing) allocation.

\begin{assumption}[Monotonicity and Convexity~\citep{marshall2013principles}]
\label{ass:convexity}
For every instance $i$, the loss function $\ell_i(m_i)$ is strictly decreasing and strictly convex with respect to the feature count $m_i$. That is:
\begin{equation}
    \ell_i'(m_i) < 0, \quad \ell_i''(m_i) > 0, \quad \forall m_i > 0.
\end{equation}
This reflects the principle of diminishing marginal returns: adding features reduces loss, but the benefit per feature decreases as the total number of features increases.

\end{assumption}

\begin{definition}[Uniform Allocation]
The Uniform strategy assigns an equal feature budget $\bar{m}$ to all instances:
\begin{equation}
    m_i^{\text{uni}} = \bar{m} = \frac{B}{N}, \quad \forall i.
\end{equation}
The total loss is denoted as $\mathcal{L}_{\text{uni}} = \sum_{i=1}^N \ell_i(\bar{m})$.
\label{uniform}
\end{definition}

\begin{definition}[Adaptive Allocation]
The Adaptive strategy seeks the optimal allocation $\{m_i^*\}_{i=1}^N$ that minimizes the total loss:
\begin{equation}
    \min_{\{m_i\}} \sum_{i=1}^N \ell_i(m_i) \quad \text{s.t.} \quad \sum_{i=1}^N m_i = B.
    \label{eq:adaptive}
\end{equation}
The total loss is denoted as $\mathcal{L}_{\text{adapt}} = \sum_{i=1}^N \ell_i(m_i^*)$.
\label{adaptive}
\end{definition}

\subsection{Optimal strategy}

\begin{theorem}
Under Assumption \ref{ass:convexity}, the adaptive strategy outperforms uniform strategy ($\mathcal{L}_{\text{adapt}} \le \mathcal{L}_{\text{uni}}$). Furthermore, the magnitude of the loss reduction is proportional to the variance in the marginal returns among the instances when resources are allocated uniformly.
\end{theorem}

\begin{proof}
We formulate the constrained optimization problem in Equation \ref{eq:adaptive} using the Lagrangian function, with $\lambda$ serving as the Lagrange multiplier for the budget constraint.
\begin{equation}
    \mathcal{L}(\bm{m}, \lambda) = \sum_{i=1}^N \ell_i(m_i) + \lambda \left( \sum_{i=1}^N m_i - B \right).
\end{equation}
The first-order optimality conditions, specifically the Karush-Kuhn-Tucker (KKT) conditions, for the resource allocation problem directly lead to the Equimarginal Principle~\citep{marshall2013principles}.
\begin{equation}
    \frac{\partial \ell_i}{\partial m_i} + \lambda = 0 \implies \ell_i'(m_i^*) = -\lambda, \quad \forall i.
\end{equation}
This implies that at the optimal allocation $m^*$, the marginal gradient is \textbf{identical} across all instances. Let this optimal gradient be $g^* = -\lambda$.
To quantify the gap $\Delta \mathcal{L} = \mathcal{L}_{\text{uni}} - \mathcal{L}_{\text{adapt}}$, we perform a second-order Taylor expansion of $\ell_i(m_i^*)$ around the uniform point $\bar{m}$:
\begin{equation}
    \ell_i(m_i^*) \approx \ell_i(\bar{m}) + \ell_i'(\bar{m})(m_i^* - \bar{m}) + \frac{1}{2}\ell_i''(\bar{m})(m_i^* - \bar{m})^2.
\end{equation}
Summing over all $i$, and noting that $\mathcal{L}_{\text{adapt}} = \sum \ell_i(m_i^*)$ and $\mathcal{L}_{\text{uni}} = \sum \ell_i(\bar{m})$:
\begin{equation}
    \mathcal{L}_{\text{adapt}} \approx \mathcal{L}_{\text{uni}} + \sum_{i=1}^N \ell_i'(\bar{m})\Delta m_i + \frac{1}{2} \sum_{i=1}^N \ell_i''(\bar{m})(\Delta m_i)^2,
\end{equation}
where $\Delta m_i = m_i^* - \bar{m}$.
From the budget constraint, we know $\sum \Delta m_i = 0$. 
However, the first-order term $\sum \ell_i'(\bar{m})\Delta m_i$ does not vanish because gradients $\ell_i'(\bar{m})$ are \textbf{heterogeneous}. 
This heterogeneity implies that not all data (tokens) require the same resource (expert budget), a conclusion empirically motivated in Figure \ref{fig:motivation} and discussed in Section \ref{sec:seqtopk}.
To relate $\Delta m_i$ to the gradients, we linearize the gradient condition. Near $\bar{m}$, we approximate:
\begin{equation}
    \ell_i'(m_i^*) \approx \ell_i'(\bar{m}) + \ell_i''(\bar{m})\Delta m_i.
\end{equation}
Since optimality requires $\ell_i'(m_i^*) = g^*$ (constant for all $i$):
\begin{equation}
    g^* \approx \ell_i'(\bar{m}) + \ell_i''(\bar{m})\Delta m_i \implies \Delta m_i \approx -\frac{\ell_i'(\bar{m}) - g^*}{\ell_i''(\bar{m})}.
\end{equation}
For analytical insight, assume a constant local curvature (Hessian) $\ell_i''(\bar{m}) \approx H > 0$ across samples. Then $\Delta m_i \approx -\frac{1}{H}(\ell_i'(\bar{m}) - g^*)$.
Substituting this back into the loss expansion :
\begin{align}
    \mathcal{L}_{\text{adapt}} - \mathcal{L}_{\text{uni}} &\approx \sum_{i=1}^N \ell_i'(\bar{m})\left[ -\frac{1}{H}(\ell_i'(\bar{m}) - g^*) \right] + \frac{1}{2} \sum_{i=1}^N H \left[ - \frac{1}{H} (\ell_i'(\bar{m}) - g^*) \right]^2  + \text{h.o.t.} \\
    &\approx -\frac{1}{H} \sum_{i=1}^N \left( \ell_i'(\bar{m})^2 - \ell_i'(\bar{m})g^* \right) +
    \frac{1}{2H} \sum_{i=1}^N \left( \ell_i'(\bar{m}) - g^* \right)^2.
\end{align}
$\sum (\ell_i'(\bar{m}) - g^*) = 0$, 
Since $\sum \Delta m_i = 0$, it implies $\sum (\ell_i'(\bar{m}) - g^*) = 0$, so $g^*$ is essentially the mean gradient $\bar{g} = \frac{1}{N}\sum \ell_i'(\bar{m})$. The expression simplifies to:
\begin{equation}
    \mathcal{L}_{\text{uni}} - \mathcal{L}_{\text{adapt}} \approx \frac{1}{2H} \sum_{i=1}^N \left( \ell_i'(\bar{m}) - \bar{g} \right)^2 \geq 0.
\end{equation}
\end{proof}

\section{Sensitivity w.r.t.~Batch Size.}
\label{sec:abl_bsz}
We further investigate the effect of evaluation batch size for different methods.
Since the total expert budget of BatchTopK is influenced by the training batch size, it is unclear whether this method would stay robust during evaluation.
As shown in \Figref{fig:batch-topk}, BatchTopK achieves peak accuracy when the training and evaluation batch sizes match (16 per device in our setup) but degrades as the evaluation batch size increases. While effective in MoE-based diffusion models, its sensitivity to batch size limits practicality, where large evaluation batches are common, leading to unstable inference.
\begin{figure}[!]
    \centering
    \includegraphics[width=0.4\linewidth]{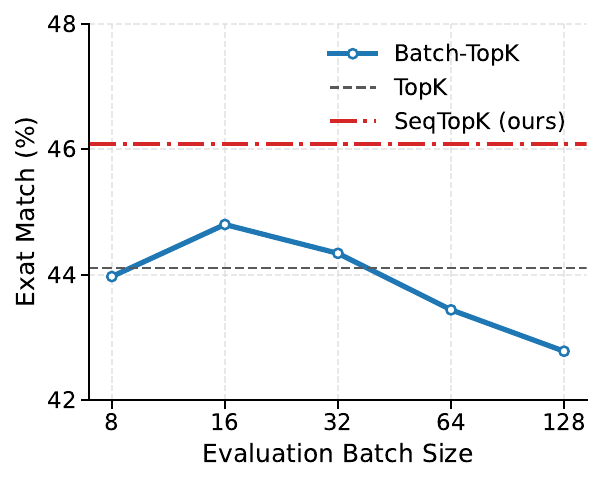}
    \caption{
    Batch-TopK's performance is highly influenced by the evaluation batch size and cannot consistently outperform TopK, while SeqTopK could.
    }
    \label{fig:batch-topk}
\end{figure}

\section{Ablations of Bound Selection}
\label{sec:ablation-upper-bond}
Table~\ref{tab:ablation-upper-bond} presents an ablation study on the selection of the upper bound for SeqTopK, evaluating configurations ranging from $K+1$ to $K+4$ and an unbounded setting across 5 benchmark datasets. 
We find that while SeqTopK’s
performance exhibits slight fluctuations depending on the specific upper bound chosen.
For example,  $K+3$ achieves the highest score on the HuamnEval dataset.
Despite these fluctuations, SeqTopK consistently delivers robust performance improvement under most of the evaluated settings.
As for the selection of lower bound, 
our core motivation, as discussed in~\Figref{fig:motivation}(a), is that not all tokens require the same computational resources.
By setting a high lower bound (e.g., $K/2$), we force the model to over-allocate computation,  artificially \textit{narrows the model's adaptive flexibility}(\emph{i.e.}, search space) and, as the Table~\ref{tab:ablation-upper-bond} shows, results in a performance drop.
\begin{table}[h]
\centering
\caption{
\textbf{Ablations of bound selection of SeqTopK}. 
We present an ablation study on the selection of the upper bound for SeqTopK, ranging from $K+1$ to $K+4$, as well as configurations with no upper bound and $K/2$ as lower bound.
}

\begin{tabular}{lccc|ccc}
\toprule

\multicolumn{3}{l}{\textit{OLMoE-A1B-7B}}
&& \textbf{GSM8k} & \textbf{MBPP} & \textbf{HumanEval} 
\\
{\textbf{Routing}}
& \multirow{2}{*}{K}
& \textbf{Lower}
& \textbf{Upper} &{0-shot} & {3-shot} & {0-shot} 
\\

\textbf{Methods} 
&
&\textbf{bound}
&\textbf{bound} & {EM} & {Pass@1} & {Pass@1} 
\\ \midrule
\addlinespace
Base
& \multirow{2}{*}{8}
& \multirow{2}{*}{K}
& \multirow{2}{*}{K}
& 15.58 
& 19.80 
& 10.97 

\\
TopK
& 
&
&
& 44.11 
& 21.04 
& 13.41 
\\
\midrule
\multirow{6}{*}{\textbf{SeqTopK}}
& \multirow{6}{*}{8}
& \multirow{6}{*}{1}
& K+1
& 46.07 
& 22.38
& 14.63
\\
& 
&
& K+2
& \colorbox{orange!10}{\textbf{46.09}}  
& \colorbox{orange!10}{\textbf{23.21}} 
& \colorbox{orange!10}{\textbf{15.24}} 
\\
& 
&
& K+3
& 45.15 
& 22.51
& \colorbox{orange!10}{\textbf{15.24}} 
\\
& 
&
& K+4
& 44.54 
& 22.00
& 15.01
\\
& 
&
& -
& 44.09 
& 21.80
& 14.02
\\
& 
& K/2
& K+2
& 44.38 
& 22.21
& 14.37
\\
\bottomrule
\end{tabular}

\label{tab:ablation-upper-bond}
\end{table}

\section{Detail Experiment Setting}
\subsection{Pre-training Experiment Setting}
\label{pre-train-setting}
We followed the training pipeline described in~\citet{wang2024remoe}, using The Pile dataset~\citep{gao2020pile} for 60,000 steps, which corresponds to approximately 30 billion tokens. 
This training duration exceeds the compute-optimal dataset size suggested by~\citet{krajewski2024scaling}, ensuring the models reach convergence.
We follow Megatron-LM~\citep{megatron-lm} by concatenating documents via $<eod>$ and slicing them into fixed-length sequences (\emph{e.g.}, 1024). 
We further employ Generalized Attention Masks and Position ID resetting to guarantee independence between packed documents.
Our pre-training experiments were conducted at two distinct sparsity levels, activating either 8 out of 64 experts or 4 out of 128 experts for each token. 
The specific configurations for each model are detailed in Table~\ref{tab:pre-train-config}.
We used a byte pair encoding (BPE) tokenizer~\citep{sennrich2015neural} and maintained consistent training parameters across all models.
We employed the AdamW optimizer~\citep{loshchilov2017decoupled} with ZeRO optimization~\citep{loshchilov2017decoupled}, using default $\beta_1=0.9$ and $\beta_2=0.999$ values. 
The learning rate was set to $5e^{-4}$
with a cosine scheduler and the coefficient for the auxiliary loss~\citep{wang2024auxiliary}is set to $0.01$.

For evaluation, we evaluate the zero-shot performance of the trained models on the following downstream tasks: LAMBADA~\citep{paperno2016lambada}; RACE~\citep{lai2017race};
ARC~\citep{clark2018think}.
All models were trained on 8 NVIDIA A100 GPUs.

\begin{table}[h]
    \centering
    \caption{Configurations for different models.}
    \label{tab:pre-train-config}
    \begin{tabular}{c|cccccccc}
        \toprule
        \textbf{Methods} & \textbf{\#Parameters} & \textbf{Total Experts} & \textbf{Activated Experts} & \textbf{hidden\_size} & \textbf{num\_layers} &\boldsymbol{$\lambda_0$} &\boldsymbol{$\alpha$}
        \\
        \midrule
        TopK & 182M & 64/128 & 8/4 & 768 & 12  &-&-\\
        ReLU & 182M & 64/128 & 8/4 & 768 & 12  &$1e^{-8}$ &1.2 \\
        SeqTopK & 182M & 64/128 & 8/4 & 768 & 12 &-&- \\
        \bottomrule
    \end{tabular}
\end{table}

\subsection{LLM fine-tuning Experiment Setting}
\label{ft-setting}
We select OLMoE-7B-A1B~\citep{muennighoff2024olmoe} and Qwen1.5-14B-A2.7B~\citep{qwen_moe} as backbone models and use the codebase provided by ~\citet{yao2025densemixer} for downstream fine-tuning.
For the math domain, we train on GSM8K~\citep{cobbe2021training} and evaluate zero-shot performance.
For the code domain, we train the
models on the Python subset of the enormous CodeAlpaca dataset~\citep{luo2023wizardcoder} to simulate a more targeted LLM customization scenario,
and evaluate on HumanEval~\citep{chen2021evaluating} and MBPP~\citep{austin2021program}.
We also incorporate evaluation on the Specialized Tasks proposed by~\citet{wang2024letexpertsticklast}.
Following standard training procedures, we pad samples to 4096 tokens. 
Sequences exceeding the context window are truncated to retain the initial segment and preserve primary context.
All models are optimized within 200 steps under the same training setup, and \textbf{identical hyperparameters}; the training hyperparameters for each model--task pair are listed in Table~\ref{tab:ft-train-config}.
In MRL-TopK, we apply a layer-wise sampling strategy where the dimension $k$ for each layer $l$ is sampled from a discrete uniform distribution over $k \sim \text{Uniform}(\{2^1, \dots, 2^{\log_2 K}\})$.
All models were trained on 8 NVIDIA A100 GPUs.

We evaluate zero-shot performance on GSM8K (Exact Match) and HumanEval (Pass@1), 3-shot performance on MBPP (Pass@1), and use GPT-4o~\citep{hurst2024gpt} to assess the quality of answers on summarization and legal tasks following ~\citet{wang2024letexpertsticklast}.

\begin{table}[h]
    \centering
    \caption{Configurations for fine-tuning MoE models}
    \begin{tabular}{cccccccc}
    \toprule
    Model  & Dataset   & Epoch &Batch\_Size &Per-GPU BS & Lr     & Warmup\_Ratio & Gradient\_Checkpointing\\ 
    \midrule
    \multirow{4}{*}{OLMOE}   
& GSM8K     
& 3     
& 256      
& 16
& 1e-6   
& 0.03          
& False                   \\
& Codealpha 
& 2     
& 256   
& 8
& 1e-6   
& 0.03          
& False                  \\ 
& Summary    
& 2     
& 256         
& 4
& 2e-5   
& 0.03          
& False                   \\                & Law       
& 4     
& 256       
& 4
&2e-5   
& 0.03          
& False                   \\ 
\midrule
\multirow{4}{*}{Qwen1.5} 
& GSM8K     
& 4     
& 384     
& 4
& 1e-6   
& 0.1           
& True                   \\  
& Codealpha 
& 2     
& 256         
& 1
& 2e-6   
& 0.1           
& True                   \\                & Summary    
& 1     
& 256     
& 2
& 1e-6 
& 0.1           
& True                   \\                & Law       
& 4     
& 128         
& 4
& 1e-5 
& 0.1           
& True                   \\ 
\bottomrule
\end{tabular}
\label{tab:ft-train-config}
\end{table}

\subsection{Discussion on Sentence Segmentation strategy}
The choice of segmentation strategy (e.g., fixed-size chunks, syntactic boundaries, or dynamic criteria) is vital for efficient large-scale training on long documents.
This often involves packing samples into one sequence for efficiency~\citep{raffel2020exploring}, which necessitates methods like the 4D attention mask proposed by~\citet{RuslanS_4D_Masks_2024} to confine attention to individual sequences.
SeqTopk could also leverage these techniques; for instance, a 4D mask could enforce syntactic boundaries for its context-level routing, keeping it within a single, semantically-consistent sentence.
\section{Fine-tuning Comparison with MoE++}
\label{sec:comp_moe++}
\begin{table}
    \centering
    \caption{\textbf{Comparison with
MoE++ (fine-tuned) on the
GSM8K benchmark.}}
    \label{tab:moe++}
    \small 
    \begin{tabular}{lccc} 
        \toprule
        \textbf{Methods} & \textbf{Training Steps} &\textbf{Acc.} \\ 
        \midrule
        Base   &0 & 15.58 \\
        TopK   &90 & 44.11 \\
        MoE++  &90 & 18.91 \\
        MoE++  &360 & 21.97 \\
        \midrule
        \textbf{SeqTopK} &90 & \textbf{46.09} \\
        \bottomrule
    \end{tabular}
\end{table}
We have added a comparison between MoE++ in a fine-tuning setting and found that SeqTopK \textbf{outperforms} MoE++ by a significant margin (\textbf{46.09} vs.\ 18.91) under equal training cost. Following~\cite{jin2024moe++}, we implemented MoE++ on top of the pre-trained OLMoE model by adding one ``zero'' and one ``copy'' expert, defined as $E_{\text{zero}}(x) = 0$ and $E_{\text{copy}}(x) = x$. As these two experts are parameter-free functions, the only newly added parameters reside in the gating network $W_g$, which expands from $\mathbb{R}^{d \times N}$ to $\mathbb{R}^{d \times (N+2)}$. 

To ensure a fair comparison and prevent the model from fighting random noise, we did not use random initialization, which we observed leads to immediate model collapse and divergence. Instead, we employed a \textit{Partial Inheritance (Warm-Start)} strategy: the original $N$ rows of the gating matrix were directly inherited from the pre-trained weights, while the two new rows for ``zero'' and ``copy'' experts were initialized using the mean statistics of the existing weights to align them with the pre-trained latent space.

As highlighted in Table~\ref{tab:moe++}, despite this alignment strategy, SeqTopK establishes a substantial {+143\%} performance advantage over MoE++. While MoE++ yields a marginal improvement over the base model, it is hindered by \textbf{fundamental architectural incompatibilities} in a fine-tuning context. Notably, even when we \textbf{extend the fine-tuning duration to 360 steps} (4$\times$ the original schedule), MoE++ only reaches an accuracy of \textit{21.97}, still failing to recover from the routing distribution shift. 
In contrast, SeqTopK integrates seamlessly with existing pre-trained checkpoints by optimizing the selection scope without modifying the router’s architecture, delivering large gains without requiring re-training from scratch.

\section{Training-Free SeqTopK}
\label{sec:training-free seqtopk}
We investigate the effect of incorporating the online SeqTopK decoding strategy into a fine-tuned TopK model \textit{without} additional training. 
As shown in Table~\ref{tab:training-free seqtopk}, the training-free SeqTopK variant with Expert Cache achieves performance comparable to TopK, highlighting its seamless compatibility with modern MoE frameworks. 
To fully exploit SeqTopK's potential, further fine-tuning (fewer than 100 steps) is required to enable the model to autonomously utilize context awareness for expert routing, as demonstrated by the results in Table~\ref{tab:performance-olmoe}.

\begin{table}[h]
\centering
\caption{\textbf{Results of \textit{Training-Free} SeqTopK on top of TopK fine-tuned model.}}

\begin{tabular}{lcc|ccccc|c}
\toprule

\multicolumn{3}{l}{\textit{OLMoE-A1B-7B}}
& \textbf{GSM8k} & \textbf{MBPP} & \textbf{HumanEval} & \textbf{Summary} & \textbf{Law}  & \textbf{Avg} \\
{\textbf{Routing}}
& \multirow{2}{*}{\textbf{K}} & \textbf{Sparsity}&{0-shot} & {3-shot} & {0-shot} & {0-shot} & \small{0-shot} &-  \\

\textbf{Methods} & &\textbf{Ratio}& {EM} & {Pass@1} & {Pass@1} & {Score} & {Score} & Score   \\ \midrule
\addlinespace
Base
& {8}
& {1/8}
& 15.58 
& 19.80 
& 10.97 
& 7.49 
& 5.70 
& 11.91 \\
\midrule
TopK
& \multirow{2}{*}{8}
& \multirow{2}{*}{1/8}
& 44.11 
& 21.04 
& 13.41 
& 45.31 
& 24.89 
& 29.74  \\
SeqTopK     
& 
& 
& 44.04      
& 20.58    
& 13.03          
& 44.87        
& 24.17    
& 29.34     \\
\bottomrule
\end{tabular}

\label{tab:training-free seqtopk}
\end{table}

\section{More Visualizations}
\label{sec:vis}
To investigate how the model handles ambiguity, we visualized the routing probability distribution for tokens that require more than $K$ experts (Figure~\ref{fig:routing_distribution}). 
As shown in Figure~\ref{fig:routing_distribution}(a), TopK exhibits a rigid, high-variance distribution with dominant spikes reaching $0.14$; this behavior is suboptimal for tokens that semantically align with multiple experts
SeqTopK addresses this by enabling a smoother probability landscape. As illustrated in Figure~\ref{fig:routing_distribution}(b), SeqTopK suppresses the sharp peaks observed in TopK (reducing the maximum probability from $0.14$ to $\approx 0.08$) and allocates more weight to secondary experts. 
This shift ensures a more balanced and uniform expert utilization, further evidenced by our visualizations of expert load variance (\Figref{fig:routing_distribution}) and expert-load histograms (\Figref{fig:expert-hist}), which provide a more intuitive perspective on the model's improved routing dynamics.

\begin{figure}[h]
    \centering

    \begin{subfigure}{1.0\textwidth}
        \centering
    \includegraphics[width=0.8\linewidth]{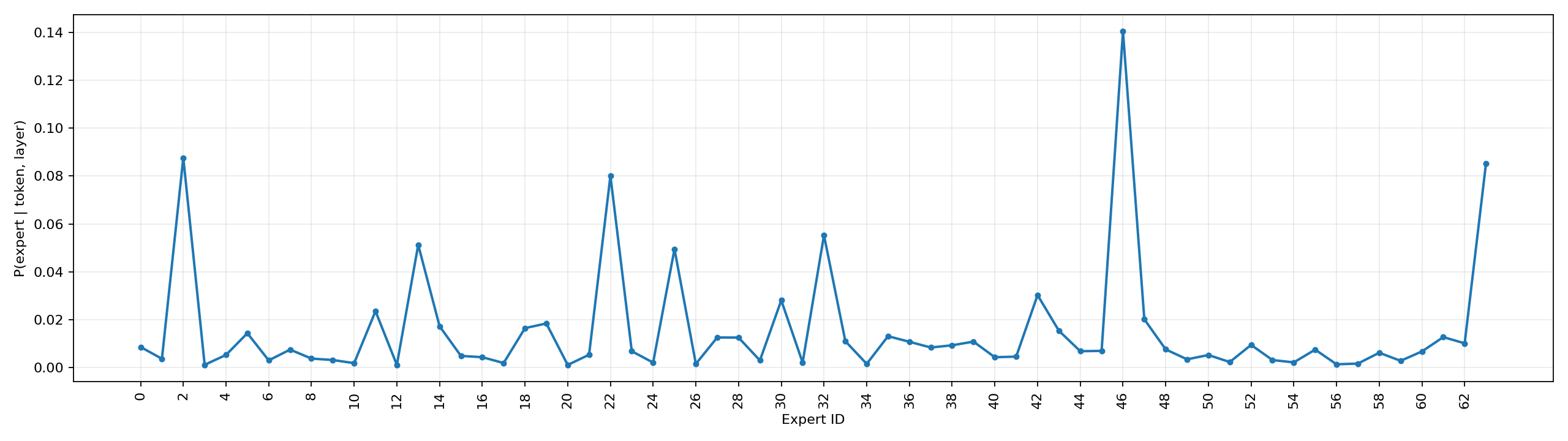} 
        \caption{TopK}
        \label{fig:top_image}
    \end{subfigure}

    \par\bigskip 
    \begin{subfigure}{1.0\textwidth}
        \centering
\includegraphics[width=0.8\linewidth]{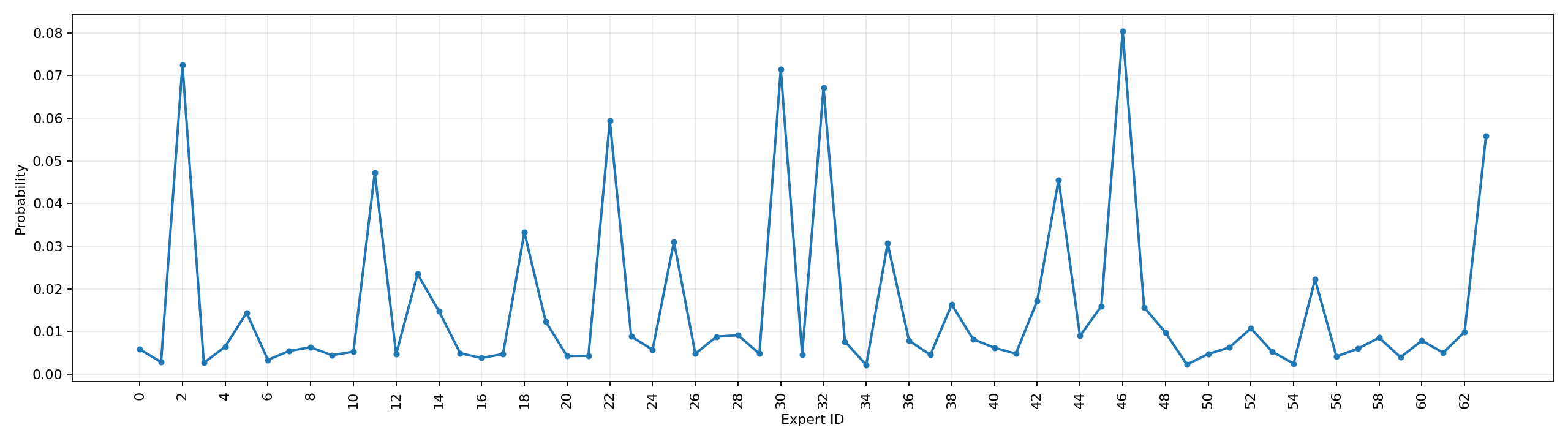}
        \caption{SeqTopK}
        \label{fig:bottom_image}
    \end{subfigure}

    \caption{ 
    \textbf{Impact of SeqTopK on Routing Distribution of token with $>K$ experts.}
    (a) \textbf{TopK} results in a highly peaked distribution where the dominant expert receives a probability of $\approx 0.14$. 
(b) \textbf{SeqTopK} enables a softer distribution for tokens requiring $>K$ experts, flattening the peak probability to $\approx 0.08$ and spreading weights more broadly across potential experts.
    }
    \label{fig:routing_distribution}
\end{figure}

\begin{figure}[t!]
    \centering
    \includegraphics[width=1\linewidth]{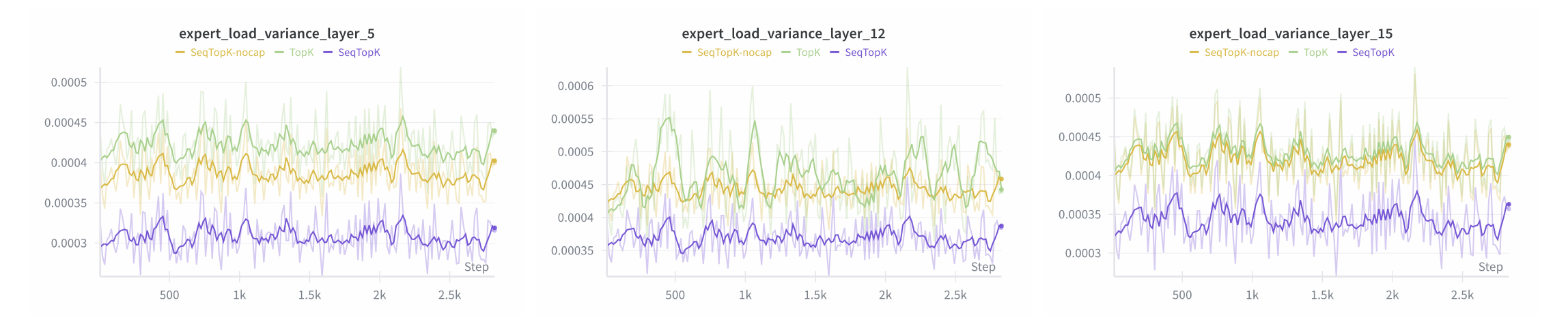}
    \caption{
\textbf{Expert load variance during training.}
We present training-time expert load variance at layers 5, 12, and 15 during OLMoE fine-tuning on GSM8K. 
SeqTopK with an upper bound consistently delivers the lowest load variance across all tested layers, ensuring highly balanced expert utilization. 
Furthermore, even the unconstrained variant of SeqTopK maintains a lower variance than the standard TopK baseline. 
    }
    \label{fig:expert-hist}
\end{figure}

\begin{figure}[t]
    \centering
\includegraphics[width=\linewidth]{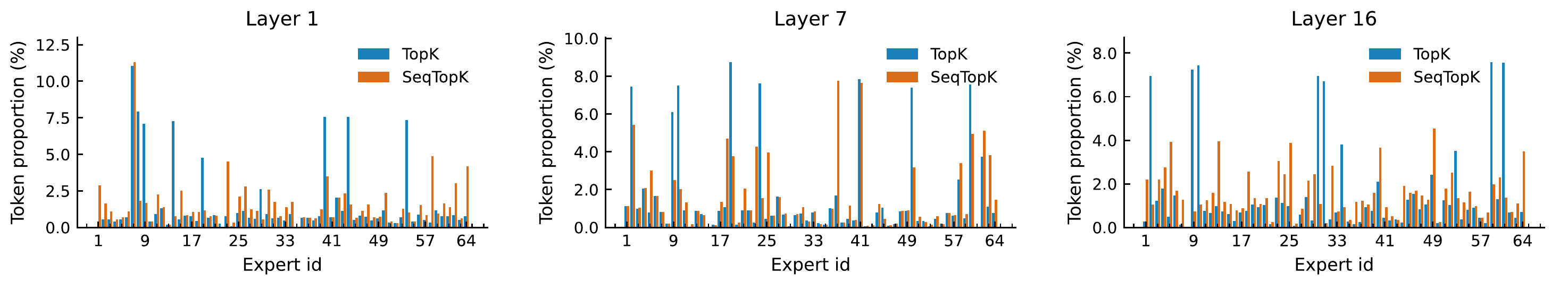}
    \caption{
\textbf{Expert-load histogram across layers.}
We present expert-load histograms at layers 1, 7, and 16, corresponding to varying entropy levels presented in \Figref{fig:router_dynamics}(b). 
SeqTopK exhibits smoother and more balanced expert utilization compared to TopK.
This optimized sequence-level allocation effectively circumvents the limitations of fixed assignment constraints, resulting in performance gains.
 }
\label{fig:normalized-entropy}
\end{figure}

\section{Limitations}
In this section, we discuss the boundaries and limitations of SeqTopK. 
First, the method is currently specialized for sparse MoE routing, which means it is inapplicable to dense architectures.
However, the core principle of \textit{adaptive budget allocation} is generalizable to other domains, such as dynamic attention pruning, which we leave it for future work.
Second, SeqTopK introduces only a slight overhead: as shown in Table~\ref{tab:inference-efficiency}, throughput decreases by $\approx 1\%$ (141.23 vs. 139.41 tokens/s), with marginal increases in memory and training time ($<1\%$). We argue that this minimal cost is a highly justified trade-off given the substantial performance gains of +5.9\% and +3.6\% demonstrated in Tables~\ref{tab:performance-olmoe} and~\ref{tab:performance-qwen}, respectively.

\end{document}